\documentclass[twocolumn]{autart}    

\usepackage{graphicx}          
\usepackage[dvips]{epsfig}    

\usepackage{cite}

\usepackage{textcomp}
\usepackage{amsmath,amsfonts,amssymb,color}

\def\endpf{\hspace*{\fill}~$\square$\par\endtrivlist\unskip}

\usepackage{epsfig}
\usepackage{psfrag}
\usepackage{algorithm}
\usepackage{algpseudocode}
\usepackage{array}
\usepackage{epstopdf}
\usepackage{mathtools}
\usepackage{enumerate}
\usepackage{soul,color}
\usepackage{wrapfig}
\usepackage{nccmath}

\newtheorem{theorem}{Theorem}

\newtheorem{lemma}{Lemma}

\newtheorem{definition}{Definition}
\newtheorem{proposition}{Proposition}
\newtheorem{assumption}{Assumption}

\newtheorem{Remark}{Remark}

\DeclareMathOperator{\diag}{diag}

\DeclarePairedDelimiter\floor{\lfloor}{\rfloor}
\DeclarePairedDelimiter{\ceil}{\lceil}{\rceil}

\begin{document}

\begin{frontmatter}

\title{Learning Linearized Models from Nonlinear Systems under Initialization Constraints with Finite Data} 

\thanks[footnoteinfo]{This work was supported by the National Science Foundation CAREER award 1653648.}

\author[1]{Lei Xin}\ead{lxinshenqing@gmail.com},  
\author[2]{Baike She}\ead{bshe6@gatech.edu}, 
\author[1]{Qi Dou}\ead{qidou@cuhk.edu.hk},
\author[3]{George Chiu}\ead{gchiu@purdue.edu},               
\author[4]{Shreyas Sundaram}\ead{sundara2\@purdue.edu},  

\address[1]{Department of Computer Science and Engineering, The Chinese University of Hong Kong, Hong Kong} 
\address[2]{School of Electrical and Computer Engineering, Georgia Institute of Technology, Atlanta, GA, 30318, USA} 
\address[3]{School of Mechanical Engineering, Purdue University, West Lafayette, IN 47907, USA}   

\address[4]{Elmore Family School of Electrical and Computer Engineering, Purdue University, West Lafayette, IN 47907, USA}



\begin{abstract}                          
The identification of a linear system model from data has wide applications in control theory. The existing work that provides finite sample guarantees for linear system identification typically uses data from a single long system trajectory under i.i.d. random inputs, and assumes that the underlying dynamics is truly linear. In contrast, we consider the problem of identifying a linearized model when the true underlying dynamics is nonlinear, given that there is a certain constraint on the region where one can initialize the experiments. We provide a multiple trajectories-based deterministic data acquisition algorithm followed by a regularized least squares algorithm, and provide a finite sample error bound on the learned linearized dynamics. Our error bound shows that one can consistently learn the linearized dynamics, and demonstrates a trade-off between the error due to nonlinearity and the error due to noise.  We validate our results through numerical experiments, where we also show the potential insufficiency of linear system identification using a single trajectory with i.i.d. random inputs, when nonlinearity does exist.
\end{abstract}

\end{frontmatter}

\section{Introduction} \label{sec: introduction}
Learning accurate predictive models from data has wide applications, including in machine learning and economics \cite{athey2018impact, mitchell2007machine}. The problem of system identification is to learn a mathematical model of a dynamical system from data. System identification is an important problem in control theory since a good model can facilitate model-based control design \cite{ljung1999system}. Although physical systems are typically nonlinear, linear models are frequently used in practice due to their simplicity \cite{rugh1996linear}, and their ability to approximate nonlinear systems around a given reference point. Consequently, it is of interest to understand identification of appropriate linear models from data generated by nonlinear systems.

Classically, theories for system identification typically focus on asymptotic aspects \cite{bauer1999consistency,jansson1998consistency}. In recent years, however, finite sample analysis for system identification has been studied extensively. The primary goal of finite sample analysis for system identification is to understand the factors that influence the error and how the error diminishes with a finite number of samples. Such analyses can also help identify system characteristics that facilitate learning and provide insights for the development of more effective algorithms. For linear system identification, existing works are either multiple trajectories-based or single trajectory-based. The multiple trajectories setup \cite{dean2019sample, fattahi2018data, zheng2020non, xin2022learning} requires the user to restart the system multiple times, with existing studies assuming that the initial state can be set to exactly zero \cite{dean2019sample, fattahi2018data, zheng2020non}. However, a major advantage of this setup is its ability to handle unstable systems. In contrast, the single trajectory setup \cite{simchowitz2018learning,oymak2019non,simchowitz2019learning,sarkar2019nonparametric,faradonbeh2018finite,sarkar2019near} performs system identification using data from a single experiment, i.e., the system does not need to reset,  but has potential risks if the system is unstable.  We note that when it comes to linear system identification, almost all existing works that have finite sample guarantees assume that the underlying system is truly linear, except for \cite{sarker2023accurate}. Furthermore, i.i.d. Gaussian inputs are typically applied to ensure persistent excitation. 

The study on nonlinear system identification is less well-understood, in general, as compared to the case for linear system identification. Recent works on finite sample analysis for nonlinear system identification include \cite{sattar2022non, mania2020active, foster2020learning}. It is worth noting that to obtain finite sample guarantees, the existing works on nonlinear system identification typically require that a certain model structure to be known in advance. However, when the specific model structure is unknown, a reasonable alternative goal is to learn a linearized model from the nonlinear system, due to the well-studied techniques on linear system control as discussed above.

There is a branch of research that studies learning a global linear system representation that completely captures the behaviours of a nonlinear system using the Koopman Operator \cite{mauroy2016linear}. In general, this approach may require carefully selected basis functions (e.g., using neural networks \cite{hao2024deep}), and the analysis focuses on the noiseless setting.  In contrast, our focus in this work is to learn a linearized system model, in the sense that the model captures the linear part of the nonlinear system after Taylor expansion around the origin, supposing that one has control over the initial conditions of the experiments. We also aim to provide finite sample guarantees when the system has noise.

Most relevant to our work are the papers \cite{sarker2023accurate, ahmadi2021safely}. The paper \cite{sarker2023accurate} provides a finite sample error bound for learning linear models from systems that have unmodeled dynamics that could capture nonlinearities, using a single system trajectory. However, the method proposed in \cite{sarker2023accurate} assumes the system dynamics is ``well-behaved'' by requiring the unmodeled dynamics/nonlinear terms to be (globally) Lipschitz \cite{cobzacs2019lipschitz}. The method also requires the system to satisfy certain additional properties to ensure consistent estimation, supposing the inputs are carefully chosen. The paper \cite{ahmadi2021safely} studies the optimal experiments initialization problem (i.e., how to optimally initialize the states of a system) for recovering the full system dynamics. On the other hand, it assumes that the underlying dynamics is noiseless. In contrast, our conference paper \cite{xin2023linearized} studies how one can learn a linearized system model from a noisy nonlinear system \cite{xin2023linearized} with arbitrarily small error without the Lipschitzness assumption, given sufficiently many short trajectories, supposing that one can arbitrarily initialize the initial conditions of the experiments.

However, we note that arbitrarily initializing system states/inputs can be challenging in practice. In contrast, sometimes one can only initialize the state-input vector within a feasible region. This could be due to physical constraints on the input, or the fact that initializing the state at certain locations is hard. Further, the paper \cite{xin2023linearized} does not provide an explicit convergence rate of the proposed system identification algorithm. 

In this paper, we address the above problems. In summary, our contributions are as follows. 
\begin{itemize}
  \item We provide a deterministic, multiple trajectories-based data acquisition algorithm, assuming one can only initialize the state-input vector within a given feasible region. Using this algorithm followed by a regularized least squares estimation algorithm, we develop a finite sample error bound of the learned linearized dynamics of a general nonlinear system. When the feasible region is an open set that contains the origin, we show that one can consistently learn the linearized dynamics with a rate of $\mathcal{O}(\frac{1}{N^{\frac{1}{4}}})$ in the worst case, where $N$ is the number of experiments. To the best of the authors' knowledge, this rate is novel in the considered setting. Our bound demonstrates a trade-off between the error due to noise and the error due to nonlinearity, and characterizes the benefits of using regularization. Our result is general in that when the system is perfectly linear, we show a learning rate that matches the existing results on learning perfectly linear systems using random inputs. When the feasible region is a convex set that does not contain the origin, we show that one can still achieve a small error given sufficiently many experiments, as long as the feasible region is not too far from the origin (which will be made clear later). 
  
  \item We provide numerical experiments to validate our results and insights, and show the potential limitation
  of linear system identification using random inputs from a single trajectory in the presence of mild nonlinearity.
  \end{itemize}
  
Our paper is organized as follows. Section \ref{notation} introduces relevant mathematical notation. Section \ref{sec:algorithm} introduces the system identification problem and the algorithms we use. In Section \ref{analysis}, we present our theoretical results. We present numerical examples in Section \ref{exp} to validate our results, and conclude in Section \ref{sec: conclusion}. The proofs are included in the appendix.

\section{Notation} \label{notation}
Vectors are taken to be column vectors unless indicated otherwise. Let $\mathbb{R}$ denote the set of real numbers. Let $\lambda_{max}(\cdot)$ and $\lambda_{min}(\cdot)$ be the largest and the smallest eigenvalue in magnitude, respectively, of a given matrix. For a given matrix $A$, we use $A'$ to denote its conjugate transpose. We use $\|A\|$, $\|A\|_{1}$ and $\|A\|_{F}$ to denote the spectral norm, $1$-norm, and Frobenius norm, respectively, of matrix $A$. We use $I_{n}$ to denote the identity matrix with dimension $n$. We use the symbol $\bmod$ to denote the modulo operation. The union of sets is denoted as $\cup$. The open $l_{1}$ ball in $d$-dimensional space with center at $x_{0}$ and radius $r$ is
denoted by $\mathcal{B}_{d}(x_{0},r)\triangleq \{x\in \mathbb{R}^{d}:\|x-x_{0}\|_{1}< r\}$. We denote $e_{i}^{d}$ as a $d$-dimensional vector with the $i$-th component equal to 1 and all other components equal to 0. The symbols $\floor{\cdot}$ and $\ceil{\cdot}$ are used to denote the floor and ceiling functions, respectively. We use $\textbf{0}$ to denote a zero vector with dimension that is
clear from the context. The symbol $\sigma(\cdot)$ is used to denote the sigma field generated by the corresponding random vectors. The symbol $\mathcal{S}^{n-1}$ is used to denote the unit sphere in $n$-dimensional space.


\section{Problem Formulation and System Identification Algorithm} \label{sec:algorithm}
Consider the following discrete time nonlinear time invariant system
\begin{equation} \label{nonlinear}
\begin{aligned}
x_{k+1}=f(z_{k})+w_{k},
\end{aligned}
\end{equation}
where $f: \mathbb{R}^{n+p} \to \mathbb{R}^{n}$, $z_{k}=\begin{bmatrix} x_{k}'&u_{k}' \end{bmatrix}^{'}\in \mathbb{R}^{n+p}$, $x_{k}\in\mathbb{R}^{n}$, $u_{k}\in \mathbb{R}^{p}$, and $w_{k}\in \mathbb{R}^{n}$. Here, $x_{k}, u_{k}$ and $w_{k}$ are the state, input, and process noise, respectively. The noise terms $w_{k}$ are  assumed to be independent sub-Gaussian random vectors with parameter $\sigma_{w}^2$, where the definition is given below \cite{rivasplata2012subgaussian}.
\begin{definition}
A real-valued random variable $w$ is called sub-Gaussian with parameter $\sigma^2$ if we have
\begin{equation*}
\begin{aligned}
&\forall \alpha\in \mathbb{R}, \mathbb{E}[\exp(\alpha{w})]\leq \exp(\frac{\alpha^2 \sigma^2}{2}).\\
\end{aligned}
\end{equation*}
A random vector $x\in \mathbb{R}^n$ is called $\sigma^2$ sub-Gaussian if for all unit vectors $v\in \mathcal{S}^{n-1}$ the random variable $v'x$ is $\sigma^2$ sub-Gaussian.
\end{definition}

Assume that for each component function of $f$, all second order partial derivatives exist and are continuous on $\mathbb{R}^{n+p}$. From Taylor's theorem \cite{courant1965introduction}, system \eqref{nonlinear} using reference point $z_{k}=\textbf{0}$ can be rewritten as
\begin{equation} \label{system}
\begin{aligned} 
x_{k+1}=A x_{k}+Bu_{k}+w_{k}+r_{k} \\
\end{aligned}
\end{equation}
when $f(\textbf{0})=\textbf{0}$,\footnote{The case for $f(\textbf{0})\neq \textbf{0}$ can be found in \cite{xin2023linearized}.} where $A\in \mathbb{R}^{n\times n},B\in \mathbb{R}^{n\times p}$, are system matrices that capture the linear part of $f(z_{k})$, and $r_{k}=h(z_{k})\in \mathbb{R}^{n}$ is a 
remainder vector that contains higher order terms that are state/input dependent, where $h: \mathbb{R}^{n+p} \to \mathbb{R}^{n}$. Note that one can study reference points other than the origin through a coordinate transformation \cite{aoki2013state}. The above model is less studied in the literature on finite sample analysis for system identification, and we consider this model in the sequel.  When the system is perfectly linear, we have $r_{k}=\textbf{0}$, which is the commonly used model in the literature. In this paper, we assume that both the state $x_{k}$ and input $u_{k}$ can be perfectly measured.
Suppose that we can restart the system multiple times from certain user-specified initial states $x_{0}$ and inputs $u_{0}$, and obtain multiple length 1 trajectories (i.e., state-input pairs obtained by running the system for a single time step, as will be explained next). Using a superscript to denote the trajectory index, we denote the set of samples we have as $\{(x^{i}_{1}, x^{i}_{0},u^{i}_{0}):1 \leq i \leq N\}$.  Our goal is to learn the linear approximation system matrices  $\Theta \triangleq \begin{bmatrix}
A&B \end{bmatrix}\in \mathbb{R}^{n\times(n+p)}$ in system \eqref{system} from the set of samples available to us.

Our result leverages the following mild assumption on the remainder vector $r_{k}=h(z_{k})$ in system \eqref{system}.
\begin{assumption} \label{ass:remainder}
Let $r_{i,k}$ denote the $i$-th component of $r_{k}$. There exist $c>0$ and $\beta=\beta(c)$ such that $|r_{i,k}|\leq \beta \|z_{k}\|_{1}^2$ for all $i \in \{1,\ldots, n\}$ and all $z_{k}\in \mathcal{B}_{n+p}(\textbf{0},c)$.  
\end{assumption}

\begin{Remark}
The above assumption is, in fact, a direct result of assuming that each component function of the original nonlinear dynamics $f$ has all second order partial derivatives being continuous on $\mathbb{R}^{n+p}$, due to Taylor's theorem for multivariable functions from \cite[Corollary~1]{folland2005higher}. Intuitively, this assumption says that the higher order terms are dominated by the second order terms, if the arguments of the function are sufficiently close to the origin. Note that it does not require the function $h$ to be  globally Lipschitz (which is the assumption used in \cite{sarker2023accurate}). As an example, consider a scalar system with the dynamics given by $f(z_{k})=x_{k}+u_{k}+x_{k}^2+x_{k}^3$. Here $r_{k}=x_{k}^2+x_{k}^3$ satisfies Assumption \ref{ass:remainder} for $c=1$ and $\beta=2$ since $|x_{k}^2+x_{k}^3|\leq |x_{k}^2|+|x_{k}^3|\leq 2|x_{k}|^2\leq 2\|z_{k}\|_{1}^2$ for all $z_{k}\in \mathcal{B}_{2}(\textbf{0},1)$, but the corresponding function $h$ is not globally Lipschitz on $\mathbb{R}^{2}$. In general, a larger $c$ may lead to a larger $\beta$. 
\end{Remark}

Let $S \subseteq \mathbb{R}^{n+p}$ be a given region that specifies where one can initialize the state/input vectors, and let $N$ be the number of experiments to perform. Let $q>0$ be a design parameter that constrains the magnitude of the initial conditions $z_{0}$. Furthermore, let $m\in \mathbb{R}^{n+p}$ be a user-specified center point parameter. We make the following assumption.

\begin{assumption} \label{ass:Safety_set}
The parameters $m$ and $q$ are chosen such that $\bar{\mathcal{B}}_{n+p}(m,q)\subseteq S$, where $\bar{\mathcal{B}}_{n+p}(m,q)\triangleq \{m+qe_{1}^{n+p}, m+qe_{2}^{n+p},\ldots, m+qe_{n+p}^{n+p}, m-qe_{1}^{n+p}, m-qe_{2}^{n+p},\ldots, m-qe_{n+p}^{n+p}\}$.  
\end{assumption}

We deploy a data collection scheme specified in Algorithm ~\ref{algo1}. 
\begin{algorithm}[H]
\caption{Data Acquisition} \label{algo1}
\textbf{Input} Number of experiments $N>0$, Norm constraint parameter $q>0$, Center point $m$ s.t. $\bar{\mathcal{B}}_{n+p}(m,q)\subseteq S$
\begin{algorithmic}[1]
\State Initialize $s_{1}=1$
\For {$i=1,\ldots, N$}
\If{$i\bmod{(n+p)}\neq 0$}
\State Set $\mathbf{q}_{i}=s_{i}\times q e_{i\bmod{(n+p)}}^{n+p}$
\State Set $z^{i}_{0}=\left[\begin{smallmatrix} x_{0}^{i'}&u_{0}^{i'} \end{smallmatrix}\right]^{'}=m+\mathbf{q}_{i}$
\State Collect $x^{i}_{1}$, where $x^{i}_{1}=Ax^{i}_{0}+Bu^{i}_{0}+w^{i}_{0}+r^{i}_{0}$
\State Set $s_{i+1}=s_{i}$
\Else
\State Set $\mathbf{q}_{i}=s_{i}\times q e_{n+p}^{n+p}$
\State Set $z^{i}_{0}=\left[\begin{smallmatrix} x_{0}^{i'}&u_{0}^{i'} \end{smallmatrix}\right]^{'}=m+\mathbf{q}_{i}$
\State Collect $x^{i}_{1}$, where $x^{i}_{1}=Ax^{i}_{0}+Bu^{i}_{0}+w^{i}_{0}+r^{i}_{0}$
\State Set $s_{i+1}=-s_{i}$ \label{sign}
\EndIf
\EndFor
\State Output  $\{(x^{i}_{1}, x^{i}_{0}, u^{i}_{0}):1 \leq i \leq N\}$
\end{algorithmic}
\end{algorithm}
\begin{Remark}
Intuitively, we want the data/initial conditions to stay as close to the origin as possible, to avoid excessive bias from the higher order terms. Hence, we may want to use a small $q$ and a small $m$ (if physical limitations allow). However, a small $q$ would lead to a small signal-to-noise ratio, which may require more samples to reduce the error. Later on in our theoretical result, we demonstrate how $q$ and $m$ will affect the finite sample estimation error bound for learning $\Theta$, and provide more details on the guidelines for selecting these parameters. The reason of using multiple length 1 trajectories is to prevent the noise from driving the system states too far from the origin, and amplifying the effects from $r_{k}$. Intuitively, the sign change in Line \ref{sign} of Algorithm \ref{algo1} ensures that the generated dataset is both “rich” and “balanced,” thus enhancing data efficiency. Technically, it also helps reduce the unwanted effects of the potentially non-zero parameter $m$. The overall idea of Algorithm \ref{algo1} is to ensure persistent excitation (i.e., the smallest eigenvalue of the sample covariance matrix becomes larger as one gets more data), subject to the constraint on bounded distance to the origin (specified by $q$ and $m$). 
\end{Remark}

Note that in applications where a simulator is being used to learn the given dynamics, it is possible to reset the system’s states and inputs to exact values. Such linearized models are important for the initial design of controllers.  However, for physical systems, resetting the initial conditions to specific values can sometimes be challenging. In Section \ref{exp}, we numerically demonstrate that the proposed identification method is robust to small perturbations in the initial states and inputs.

We establish some definitions below. Define the batch matrices 
\begin{equation} 
\begin{aligned} 
&X=\begin{bmatrix} x^{1}_{1}&x^{2}_{1}&\cdots& x^{N}_{1}\end{bmatrix}\in \mathbb{R}^{n\times N} \\
&W=\begin{bmatrix} w^{1}_{0}&w^{2}_{0}&\cdots& w^{N}_{0}\end{bmatrix}\in \mathbb{R}^{n \times N}\\
&R=\begin{bmatrix} r^{1}_{0}&r^{2}_{0}&\cdots& r^{N}_{0}\end{bmatrix}\in \mathbb{R}^{n \times N}\\
&Z=\begin{bmatrix} z^{1}_{0}&z^{2}_{0}&\cdots &z^{N}_{0}\end{bmatrix}\in \mathbb{R}^{(n+p)\times N}.
\end{aligned}
\end{equation}

Recalling that $\Theta = \begin{bmatrix}
A&B \end{bmatrix}$, we have the relationship
\begin{equation}
\begin{aligned} 
X=\Theta Z+W+R. \\
\end{aligned}
\end{equation}

To learn the linear model $\Theta$, we would like to solve the following regularized least squares problem
\begin{equation*}
\begin{aligned}
  \mathop{\min}_{\tilde{\Theta}\in \mathbb{R}^{n\times (n+p)}} \{\|X-\tilde{\Theta}Z\|^{2}_{F}+\lambda \|\tilde{\Theta}\|^2_{F}\},
\end{aligned}
\end{equation*}
where $\lambda\geq 0$ is a regularization parameter. The closed-form solution of the above problem is given by
\begin{equation} 
\begin{aligned}
\hat{\Theta}=XZ^{'}(ZZ'+\lambda I_{n+p})^{-1},
\end{aligned}
\end{equation}
under the invertibility assumption \cite{hoerl1970ridge}. The estimation error is then given by
\begin{equation} 
\begin{aligned}
\|\hat{\Theta}-\Theta\|&=\|-\lambda\Theta(ZZ'+\lambda I_{n+p})^{-1}\\
&+WZ'(ZZ'+\lambda I_{n+p})^{-1}\\
&+RZ'(ZZ'+\lambda I_{n+p})^{-1}\| .\label{error}
\end{aligned}
\end{equation}

For the ease of reference, the above steps are encapsulated in Algorithm \ref{algo2}.

\begin{algorithm}[H]
\caption{System Identification Using Multiple Length $1$ Trajectories} \label{algo2}
\textbf{Input} Dataset $\{(x^{i}_{1}, x^{i}_{0}, u^{i}_{0}):1 \leq i \leq N\}$, regularization parameter $\lambda\geq 0$
\begin{algorithmic}[1]  
\State Construct the matrices $X,Z$. Compute $\hat{\Theta}=XZ'(ZZ'+\lambda I_{n+p})^{-1}$.
\State Extract the estimated system matrices $A,B$ from the estimate $\hat{\Theta}=\begin{bmatrix}\hat{A}&\hat{B}\end{bmatrix}$.
\end{algorithmic}
\end{algorithm}

In the next section, we provide a finite sample bound of the system identification error \eqref{error} using Algorithm \ref{algo1} and Algorithm \ref{algo2}. The bound explicitly characterizes how the error depends on $N$, $q$, $\sigma_{w}$, $\beta$, $\lambda$, and other system parameters, and will provide guidance on selecting $q,\lambda$.
 

\section{Theoretical Analysis} \label{analysis}
We provide some intermediate results first in Section \ref{intermediate}; the proofs can be found in the Appendix. Our main results are presented in Section \ref{main}.
\subsection{Intermediate results} \label{intermediate}
The following result shows the persistent excitation property of the data acquisition algorithm  (Algorithm \ref{algo1}). 
\begin{lemma} 
Suppose that Algorithm \ref{algo1} is used to generate data, and Assumption \ref{ass:Safety_set} holds. Let $N\geq 4(n+p)$. Then we have the following inequalities
\begin{equation*} 
\begin{aligned} 
&\lambda_{min}(ZZ')\geq \frac{Nq^2}{2(n+p)},\\
&\lambda_{max}(ZZ')\leq N(2\|m\|^2+\frac{2q^2}{n+p}).
\end{aligned}
\end{equation*}

\label{lemma:PE}
\end{lemma}


We have the following upper bound for the contribution due to the noise terms.
\begin{lemma} \label{bound noise}
Suppose that Algorithm \ref{algo1} is used to generate data, and Assumption \ref{ass:Safety_set} holds. Let $N\geq 4(n+p)$. Then for any fixed $\delta \in (0,1)$,  we have with probability at least $1-\delta$
\begin{equation*}
\begin{aligned}
&\|WZ'(ZZ'+\lambda I_{n+p})^{-1/2}\|\\
&\leq 3 \sigma_{w} \sqrt{\log\frac{9^n}{\delta}+(n+p)\log(1+\frac{4\|m\|^2(n+p)+4q^2}{q^2+\zeta})},
\end{aligned}
\end{equation*}
where $\zeta=\frac{4\lambda(n+p)}{N}$.
\end{lemma}

Next, we bound the contribution from the higher order terms.
\begin{lemma} \label{bound nonlinearity}
Suppose that Algorithm \ref{algo1} is used to generate data, and Assumption \ref{ass:Safety_set} holds. Let $N\geq 4(n+p)$. Fix constants $c$ and $\beta$ that satisfy Assumption \ref{ass:remainder}, and denote $\gamma=\frac{\lambda(n+p)}{Nq^2}$. Then if $\|m\|_{1}\leq (\sqrt{b}-1)q$ for some constant $b>0$ and $\|m\|_{1}+q< c$, we have 
\begin{equation} 
\begin{aligned}
&\|RZ'(ZZ'+\lambda I_{n+p})^{-1}\|\\
&\leq \sqrt{\frac{2\beta^2(n^2+np)}{1+\gamma}}bq+\frac{2 (n+p)\sqrt{\lambda Nn\beta^{2} b^2q^4}}{Nq^2+2\lambda (n+p)}.
\end{aligned}
\end{equation}
\end{lemma}

\subsection{Main Results} \label{main}
Now we present our main theoretical result, a finite sample upper bound of the system identification error \eqref{error}.
\begin{theorem} \label{thm1}
Suppose that Algorithm \ref{algo1} is used to generate data, and Assumption \ref{ass:Safety_set} holds. Let $N\geq 4(n+p)$. Fix constants $c$ and $\beta$ that satisfy Assumption \ref{ass:remainder}, and a confidence parameter $\delta \in (0,1)$. Then if $\|m\|_{1}\leq (\sqrt{b}-1)q$ for some constant $b>0$ and $\|m\|_{1}+q< c$, with probability at least $1-\delta$, the estimation error of Algorithm \ref{algo2} satisfies
\begin{equation} \label{thm1_b}
\begin{aligned}
&\|\hat{\Theta}-\Theta\|\\
&\leq \underbrace{\frac{5 \sigma_{w} \sqrt{\log\frac{9^n}{\delta}+(n+p)\log(1+\frac{4\|m\|^2(n+p)+4q^2}{q^2})}}{\sqrt{Nq^2/(n+p)+\lambda}}}_\text{Error due to noise}\\
&+\underbrace{\sqrt{\frac{2(n^2+np)}{1+\gamma}}\beta b q}_\text{Error due to nonlinearity}\\
&+\underbrace{\frac{2(n+p)(\lambda\|\Theta\|+\sqrt{\lambda Nn\beta^2 b^2q^4})}{2\lambda(n+p)+Nq^2}}_\text{Error due to regularization},
\end{aligned}
\end{equation}
where $\gamma=\frac{\lambda(n+p)}{Nq^2}$.
\end{theorem}
\begin{pf}
Recall the estimation error in \eqref{error}. We have
\begin{equation} 
\begin{aligned}
\|\hat{\Theta}-\Theta\|&\leq\lambda\|\Theta\|\|(ZZ'+\lambda I_{n+p})^{-1}\|\\
&+\|RZ'(ZZ'+\lambda I_{n+p})^{-1}\|\\
&+\|WZ'(ZZ'+\lambda I_{n+p})^{-1/2}\| \times\\
&\|(ZZ'+\lambda I_{n+p})^{-1/2}\|.\\
\end{aligned}
\end{equation}
Noting that 
\begin{equation} 
\begin{aligned}
\|(ZZ'+\lambda I_{n+p})^{-1/2}\|&=\frac{1}{\sqrt{\lambda_{min}(ZZ'+\lambda I_{n+p})}}\\
&= \frac{1}{\sqrt{\lambda_{min}(ZZ')+\lambda}},
\end{aligned}
\end{equation}
the result directly follows from applying Lemma \ref{lemma:PE}, Lemma \ref{bound noise}, and Lemma \ref{bound nonlinearity} after some algebraic manipulations.
\end{pf}
\begin{Remark} 
In practice, the parameters (or their upper bounds) in the bound of Theorem \ref{thm1} can be obtained from prior knowledge and/or from similar systems with known models. Note that Theorem \ref{thm1} holds irrespective of the spectral radius of the system matrix $A$, which captures a key advantage of the multiple trajectories setup. Additionally, the error bound is non-zero with finite data (and other parameters of the algorithms) when noise is present in the system. The requirement of a minimum $N$ can be treated as a burn-in time, which is common in the literature \cite{dean2019sample, simchowitz2018learning}.  Below we discuss other key insights provided by Theorem \ref{thm1}.

\textbf{Convergence rates for truly linear systems:}
Suppose that $\lambda=0$. Further, suppose that the feasible region $S$ is the entire $\mathbb{R}^{n+p}$. In such case, one can pick $m$ to be the origin, and set $b=1$. When the system is perfectly linear, one has $\beta=0$. Consequently, the upper bound in Theorem \ref{thm1} only contains the error due to noise, which goes to zero with a rate of $\mathcal{O}(\frac{1}{\sqrt{N}})$. This implies that our algorithm achieves a convergence rate comparable to the results in the existing literature for learning perfectly linear systems using random inputs \cite{dean2019sample,sarkar2019near}. Further, the error also converges to zero with a rate of $\mathcal{O}(\frac{1}{q})$. This captures the intuition that a larger signal-to-noise ratio is helpful for learning. 

\textbf{Trade-off between error due to noise and error due to nonlinearity:} Suppose that $\lambda=0$. Further, suppose that the feasible region $S$ is an open set that contains the origin. To make the error bound smaller, one can again pick $m$ to be the origin and set $b=1$. 
When nonlinearity does exist, i.e., $\beta >0$, one can observe that the error due to nonlinearity scales linearly with respect to $\beta$. This error can be made arbitrarily small by choosing a smaller $q$ in Algorithm \ref{algo1} (where $q$ captures the magnitude of the initial conditions when $m=\textbf{0}$), due to the linear dependence of $q$ on the second term of the error bound. On the other hand, a smaller $q$ would also make the denominator of the term capturing error due to noise small. Intuitively, a smaller $q$ corresponds to a smaller signal-to-noise ratio, which leads to a larger error due to noise. In other words, if one picks initial conditions that are close enough to the reference point (by setting $q$ to be small), one would have less bias due to nonlinearity, at the cost of having a smaller signal-to-noise ratio (thus a larger error due to noise). However, the error due to noise can be decreased by increasing the number of experiments $N$.

Although optimally balancing the trade-off between error due to noise and error due to nonlinearity can be challenging, general guidelines can be provided based on the bound in Theorem \ref{thm1}. Specifically, if one can afford to generate a large amount of data, it is preferable to use a small $q$ due to the low bias introduced by the nonlinear terms, and the small error introduced by the noise (which is due to the large amount of data). In contrast, if one can only generate a limited amount of data, a larger $q$ can be more beneficial, especially when the noise is large (i.e., $\sigma_w$ is large). These insights are different from system identification for truly linear systems. Asymptotically, one can set $q=\frac{c_{0}}{N^{\frac{1}{4}}}$ for some positive constant $c_{0}$ to achieve consistency, where the convergence rate is then given by $\mathcal{O}(\frac{1}{N^{\frac{1}{4}}})$.

\textbf{Effect of the feasible region:}
Suppose that $\lambda=0, \beta\neq 0$. When $S$ is a convex set that does not contain the origin, one cannot set $m=\textbf{0}$ and $b=1$ to satisfy the condition $\|m\|_{1}\leq (\sqrt{b}-1)q$ for arbitrary $q$. In such case, if $m$ is chosen to be far from the origin, a larger $b$ is required for a fixed $q$, i.e., there has to be a larger error due to nonlinearity. Hence, one may want to pick a point $m$ with the smallest possible norm (subject to the constraint $\bar{\mathcal{B}}_{n+p}(m,q)\subseteq S$). When $m$ is close to the origin (i.e., $\|m\|_{1}$ is small), one can pick $b,q$ to be small such that the error due to nonlinearity is small. One can then decrease the error due to noise using a large amount of samples $N$ to make the overall error small.

\textbf{Benefits of regularization:}
Suppose that $\beta\neq 0$ and $m$, $N$, and $q$ are fixed. As $\lambda$ increases, we observe that both the error due to noise and the error due to nonlinearity approach zero, and the error due to regularization converges to $\|\Theta\|$. Consequently, a general guideline for setting $\lambda$ is to choose a large value if 1) $\sigma_{w}$ is large (the system is very noisy), 2) $\beta$ is large (the system has strong nonlinearity), and/or 3) $b$ is large (the feasible region is far from the origin), while $\|\Theta\|$ is small. In this case, the error bound is dominated by the third term (error due to regularization), which is small because $\|\Theta\|$ is small. However, obtaining the optimal $\lambda$ is challenging if (some upper bounds of) the parameters in \eqref{thm1_b} are unknown in advance. In practice, one may try various values of $\lambda$ from a given range (e.g., from 0 to 10) and leverage cross-validation techniques \cite{refaeilzadeh2009cross} to select a good value of $\lambda$. We also demonstrate this approach in Section \ref{exp}.
 \end{Remark}

Theorem \ref{thm1} captures the accuracy of the learned linearized model. The following result provides a bound on the error in state prediction between the learned model and the actual nonlinear function $f$.
\begin{proposition}
Fix constants $c$ and $\beta$ that satisfy Assumption \ref{ass:remainder}, and consider a fixed $z_k \in \mathcal{B}_{n+p}(\textbf{0},c)$. The state prediction using the learned model $\hat{\Theta}$ satisfies
\begin{equation*}
\begin{aligned}
\|\hat{\Theta}z_{k}-f(z_k)\| \leq \|\hat{\Theta}-\Theta\|\|z_k\|+\sqrt{n} \beta \|z_k\|^2_{1}.
\end{aligned}
\end{equation*}
\end{proposition}
\begin{pf}
We have
\begin{equation*}
\begin{aligned}
\|\hat{\Theta}z_{k}-f(z_k)\|&= \|\hat{\Theta}z_{k}-\Theta z_k-r_k\|\\
&\leq\|\hat{\Theta}-\Theta\|\|z_k\|+\|r_k\|\\
& \leq \|\hat{\Theta}-\Theta\|\|z_k\|+\sqrt{n} \beta \|z_k\|^2_{1}, \\
\end{aligned}
\end{equation*}
where we used the inequality that $\|r_{k}\|\leq \sqrt{n} \max\limits_{i=1,\ldots,n}|r_{i,k}|$ and Assumption \ref{ass:remainder} in the last inequality.
\end{pf}

The above result states that the state prediction using the learned linear model is close to the output of the actual nonlinear function if the learned model is accurate and the state/input vector remains close to the origin. Note that the second term in the error bound goes to zero faster than the first term as the norm of $z_{k}$ decreases. This implies that the prediction error is essentially dominated by the accuracy of the learned model for small norms of $z_{k}$.


\section{Numerical Examples} \label{exp}
In this section, we provide simulated numerical examples to validate the insights for system identification using Algorithms \ref{algo1} and~\ref{algo2}. We also compare the results against the single trajectory setup, where the input is set to be independent zero mean Gaussian. More specifically, we still use Algorithm \ref{algo2} in the single trajectory setup, but the dataset is generated without restarting the system, see \cite{sarkar2019near, ye2021sample} for examples. Such comparisons are made since Gaussian inputs are commonly used in the literature on linear system identification \cite{dean2019sample,oymak2019non}. For simplicity, we set $\lambda=0$ for all experiments. All results are averaged over 100independent experiments.
\subsection{System with mild nonlinearity and $m=\textbf{0}$} \label{pend}
In the first example, we investigate the performance of the system identification algorithms under mild nonlinearity. The model we consider is given by 
\begin{equation} 
\begin{aligned}
\begin{bmatrix} 
x_{1,k+1}\\
x_{2,k+1}\\
\end{bmatrix}=
\begin{bmatrix} 
x_{1,k}+0.1x_{2,k}\\
-0.98\sin(x_{1,k})+x_{2,k}+0.1u_{k}\\
\end{bmatrix}+w_{k},
\end{aligned}
\end{equation}
which is obtained by discretizing a nonlinear pendulum using Euler's method.\footnote{https://courses.engr.illinois.edu/ece486/fa2019/handbook/\\lec02.html} We set $w_{k}$ to be independent Gaussian random vectors with zero mean and covariance matrix given by $0.25 I_{2}$. The linearized system matrices around the origin are given by 
\begin{equation}
\begin{aligned}
A=
\begin{bmatrix}
1&0.1&\\
-0.98&1\\
\end{bmatrix},
B=
\begin{bmatrix}
0\\
0.1\\
\end{bmatrix}.
\end{aligned}
\end{equation}
It can be verified that $r_{k}=\begin{bmatrix}0&-0.98\sin(x_{1,k})+0.98 x_{1,k}\end{bmatrix}'$ satisfies Assumption \ref{ass:remainder} with $\beta=1$ and $c=2$.

\begin{figure}[ht]
\minipage[t]{0.43\textwidth}
\includegraphics[width=\linewidth]{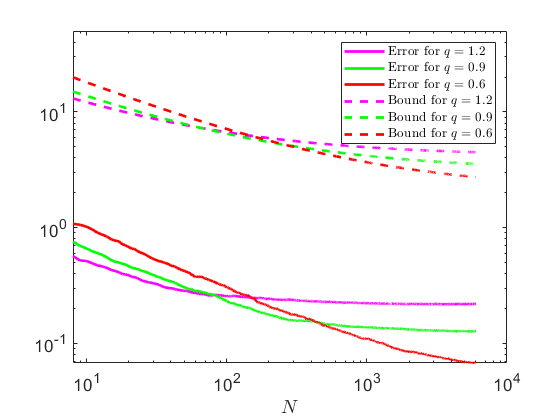}
\caption{System identification error and bound with different $q$, mild nonlinearity}
\label{pend_multi}
\endminipage \hfill
\end{figure}

We plot the system identification error using Algorithms~\ref{algo1} and~\ref{algo2} versus the number of experiments $N$ for $q=1.2$, $0.9$, and $0.6$ in Fig.~\ref{pend_multi}. We also plot the bounds in Theorem \ref{thm1} with $\delta=0.1$. As can be observed, a smaller $q$ can lead to a larger overall error when $N$ is small (i.e., when there is only a small amount of data) due to the significant error caused by noise. However, a smaller $q$ may eventually result in a smaller overall error when $N$ becomes large enough. In other words, with a large amount of data, the error due to noise diminishes, leaving only the error due to nonlinearity, which is small for small $q$. This confirms our findings in Theorem \ref{thm1}.

\begin{figure}[ht] 
\minipage[t]{0.43\textwidth}
\includegraphics[width=\linewidth]{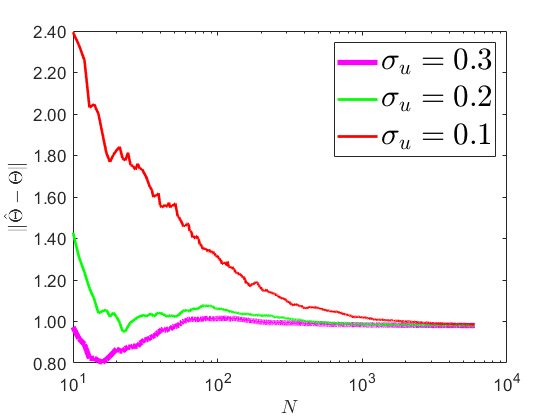}
\caption{System identification error using a single trajectory with different $\sigma_{u}$, mild nonlinearity}
\label{pend_sing}
\endminipage \hfill
\end{figure}

In the single trajectory setup, we plot the error using i.i.d zero mean Gaussian inputs with different variance $\sigma_{u}^2$, where $N$ here represents the number of samples used in the single trajectory. The initial state is set to zero. A common heuristic is that one should apply small inputs to learn a good linear approximation around a given reference point, i.e., the variance $\sigma_{u}^2$ should be small. However, as shown in Fig.~\ref{pend_sing}, the error plateaus at around 1, even for small variance inputs. The key reason is that the random input and process noise can always drive the system states to undesired regions and excite the higher order terms, unless the input is carefully designed. In fact, the paper \cite{sarker2023accurate} shows that random inputs in the single trajectory setup could result in inconsistent estimation under certain conditions even for Lipschitz nonlinearity. 

\begin{figure}[ht]
\minipage[t]{0.43\textwidth}
\includegraphics[width=\linewidth]{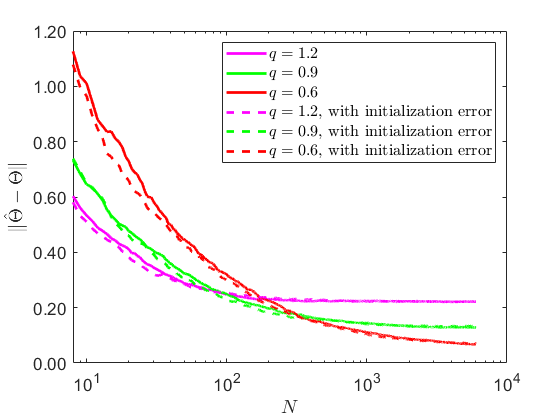}
\caption{System identification error under initialization error}
\label{In_error}
\endminipage \hfill
\end{figure}

Next, to capture scenarios where setting initial conditions to exact values is difficult, we test the robustness of the algorithms under small initialization errors. In Fig.~\ref{In_error}, we plot the system identification error under initialization errors with different values of $q$. Specifically, we add small zero-mean i.i.d. Gaussian noise to the data generated by Algorithm \ref{algo1}, where the covariance matrix is set to $0.1^2 I_3$. As can be seen from Fig.~\ref{In_error}, the small perturbations added to the dataset have negligible effects on the system identification error, demonstrating that the algorithms are robust to small perturbations. However, we conjecture that the smallest achievable error depends on the magnitude of the covariance matrix of the initialization error. Intuitively, if the noise is large with high probability, staying close to the origin becomes difficult, leading to increased error due to nonlinearity. We leave a detailed analysis for future work.

\subsection{System with strong nonlinearity and $m\neq \textbf{0}$} \label{strong}
In the second example, we investigate the performance of the system identification algorithms under strong nonlinearity (where the assumption of Lipschitzness used in \cite{sarker2023accurate} no longer holds). Further, we assume that $m\neq \textbf{0}$, which could happen if the feasible region is a convex set that does not contain the origin. The model we consider here is given by

\begin{equation} 
\begin{aligned}
\begin{bmatrix} 
x_{1,k+1}\\
x_{2,k+1}\\
\end{bmatrix}
&=\begin{bmatrix} 
0.9 & 0.6 \\
0 & 0.8 \\
\end{bmatrix}
\begin{bmatrix} 
x_{1,k}\\
x_{2,k}\\
\end{bmatrix}
+\begin{bmatrix} 
1\\
1\\
\end{bmatrix}u_{k}\\
&+\begin{bmatrix} 
x_{1,k}^3+x_{2,k}^5\\
x_{1,k}x_{2,k}^2\\
\end{bmatrix}
+w_{k},
\end{aligned}
\end{equation}
where we again set $w_{k}$ to be independent Gaussian random vectors with zero mean and covariance matrix given by $0.1^2 I_{2}$. 

\begin{figure}[ht] 
\minipage[t]{0.43\textwidth}
\includegraphics[width=\linewidth]{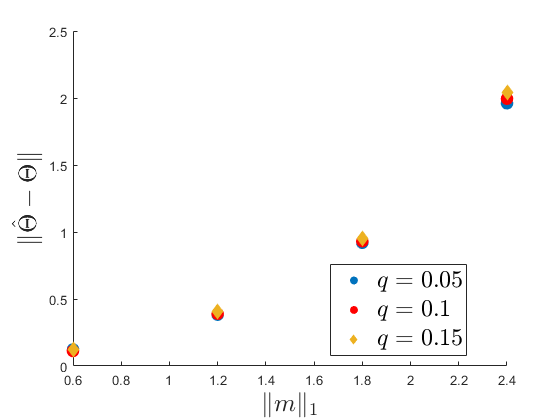}
\caption{System identification error using Algorithms \ref{algo1}-\ref{algo2} with different $m,q$. $N=10000$, strong nonlinearity}
\label{strong_multi}
\endminipage \hfill
\end{figure}

We plot the system identification error in Fig.~\ref{strong_multi} using Algorithms~\ref{algo1} and~\ref{algo2} with $N=10,000$. We set $m=\begin{bmatrix} 0.2&0.2&0.2\\ \end{bmatrix}', \begin{bmatrix} 0.4&0.4&0.4\\ \end{bmatrix}', \begin{bmatrix} 0.6&0.6&0.6\\ \end{bmatrix}'$, $\begin{bmatrix} 1.2&1.2&1.2\\ \end{bmatrix}'$ and $q=0.05, 0.1,0.15$ in these experiments. Since $N$ is sufficiently large, the errors presented here are almost purely corresponding to the error due to nonlinearity.  As can be observed, for fixed values of $q$, a larger $\|m\|_{1}$ implies a larger error due to nonlinearity, which corresponds to a larger overall error under large amount of samples. This implies that it is important to choose the center point $m$ to be close to the origin, subject to the constraint specified by the feasible region $S$.  Furthermore, for a fixed $m$, we see that a smaller $q$ could result in a smaller error when $N$ is large. These insights are consistent with our observations in Theorem \ref{thm1}.

\subsection{Effects of regularization}
\begin{figure}[ht] 
\minipage[t]{0.43\textwidth}
\includegraphics[width=\linewidth]{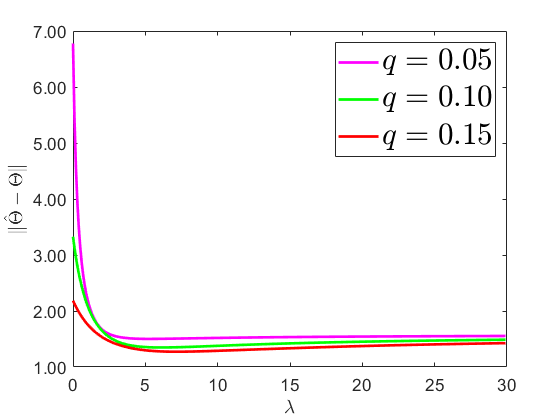}
\caption{System identification error using different regularization parameter $\lambda$}
\label{Reg}
\endminipage \hfill
\end{figure}

We consider the same system as in Section \ref{pend}, where the covariance matrix is set to $4I_3$. We set $N=500$ and plot the error for $q=0.05$, $0.1$, and $0.15$ with different values of $\lambda$, where the increment is 0.1.  As can be observed in Fig.~\ref{Reg}, a non-zero regularization parameter $\lambda$ helps reduce the error. Indeed, as discussed in Theorem~\ref{thm1}, a relatively large $\lambda$ is particularly beneficial when the nonlinear system is subject to strong noise.

We also use 10-fold cross-validation \cite{manorathna2020k} to illustrate the empirical selection of an appropriate regularization parameter $\lambda$. Specifically, we evenly split the dataset into 10 subsets. Fixing $\lambda$, for each subset, we compute the norm of the prediction error using the model learned from the remaining 9 subsets. We then average the prediction errors to obtain a performance metric for that fixed $\lambda$. Finally, we repeat this procedure across all candidate values of $\lambda$, and select the $\lambda$ that gives the best performance metric (i.e., the smallest average prediction error).  The optimal $\lambda$ obtained using the above procedure is 15.8, 14.5, and 18.5 (on average) for  $q=0.05, 0.1$, and $0.15$, respectively. Although these values do not exactly align with the true optimal $\lambda$ due to noise and the fact that the prediction error is only a proxy for the norm error considered in this paper, they demonstrate
the benefit of leveraging a non-zero regularization parameter $\lambda$.

\section{Conclusion} \label{sec: conclusion}
In this paper, we proposed a data acquisition algorithm followed by a regularized least squares algorithm to learn the linearized model of a system. Unlike existing works, we assume that the underlying dynamics could be nonlinear. We presented a finite sample error bound of the algorithms. When the feasible region for experiments initialization is an open set that contains the origin,  our bound shows that one can learn the linearized dynamics with a rate of $\mathcal{O}(\frac{1}{N^{\frac{1}{4}}})$, and demonstrates a trade-off between the error due to noise and the error due to nonlinearity. When the feasible region is a convex set that does not contain the origin, we show that one can still achieve a small error, provided that the region is not too far from the origin and sufficient samples are available. In future work, we will focus on developing algorithms with improved sample efficiency that require less physical precision in experimental hardware for setting initial conditions. Additionally, investigating the effects of measurement noise will be another potential direction for further research.

\bibliographystyle{plain}        
\bibliography{autosam}    
\section{Appendix} 
\subsection{Auxiliary Results}
\begin{lemma}(\cite[Lemma~5]{xin2023learning}) \label{martingale_bound_multi}
Let $\{\mathcal{F}_{t}\}_{t\geq 0}$ be a filtration. Let $\{{w}_{t}\}_{t\geq 1}$ be a  $\mathbb{R}^{n}$-valued stochastic process such that $w_{t}$ is $\mathcal{F}_{t}$-measurable, and $w_{t}$ is conditionally sub-Gaussian on $\mathcal{F}_{t-1}$ with parameter $R^2$. Let $\{z_{t}\}_{t\geq 1}$ be a $\mathbb{R}^{m}$-valued stochastic process such that $z_{t}$
is $\mathcal{F}_{t-1}$-measurable. Assume that $V$ is a $m\times m$ dimensional positive definite matrix. For all $t\geq 0$, define
\begin{equation*}
\begin{aligned}
&\bar{V}_{t}=V+\sum_{s=1}^{t}z_{s}z_{s}', S_{t}=\sum_{s=1}^{t}z_{s}w_{s}'.\\
\end{aligned}
\end{equation*}
Then, for any $\delta\in (0,1)$, and for all $t\geq0$,
\begin{equation*}
\begin{aligned}
&P(\|\bar{V}_{t}^{-\frac{1}{2}}S_{t}\|\leq\sqrt{\frac{32}{9}R^{2}(\log\frac{9^n}{\delta}+\frac{1}{2}\log\det(\bar{V}_{t}V^{-1})})\\
&\geq 1-\delta.\\
\end{aligned}
\end{equation*}
\end{lemma}

\begin{lemma} (\cite[Lemma~3]{chan1985hermitian}) \label{inverse bound}
Let $A\in \mathbb{R}^{n\times n}$ and $B\in \mathbb{R}^{n\times n}$ be positive definite matrices. If $A \preceq B$, then we have $A^{-1}\succeq B^{-1}$.
\end{lemma}

\begin{lemma} (\cite[Lemma~11]{xin2023learning})\label{norm bound}
Let $A\in \mathbb{R}^{n\times n}$ and $B\in \mathbb{R}^{n\times n}$ be positive semidefinite matrices. Let $C \in \mathbb{R}^{n\times m}$. If $A \preceq B$, then we have
\begin{equation*}
\begin{aligned}
\|A^{\frac{1}{2}}C\|\leq \|B^{\frac{1}{2}}C\|.
\end{aligned}
\end{equation*}
\end{lemma}

\subsection{Proof of Lemma \ref{lemma:PE}}
To ease the notation, we write $e_{i}^{n+p}$ as $e_{i}$ for $i=1,\ldots,n+p$ in the sequel. We focus on the lower bound first. Denote $N_{1}=\floor{\frac{N}{2(n+p)}}\times {2(n+p)}$. Since the assumption $N\geq 4(n+p)$ implies $N_{1}\geq 4(n+p)$, we have
\begin{equation} \label{3 terms lower}
\begin{aligned} 
ZZ'&=\sum_{i=1}^{N}z^{i}_{0}z^{i'}_{0}\succeq\sum_{i=1}^{N_{1}}z^{i}_{0}z^{i'}_{0}=\sum_{i=1}^{N_{1}}(m+\mathbf{q}_{i})(m+\mathbf{q}_{i})'\\
&=\sum_{i=1}^{N_{1}}\mathbf{q}_{i}\mathbf{q}_{i}'+\sum_{i=1}^{N_{1}}m\mathbf{q}_{i}'+\sum_{i=1}^{N_{1}}\mathbf{q}_{i}m'+N_{1}mm'.\\
\end{aligned}
\end{equation}

For the first summation after the last equality in \eqref{3 terms lower}, we have
\begin{equation} 
\begin{aligned} 
\sum_{i=1}^{N_{1}}\mathbf{q}_{i}\mathbf{q}_{i}'=&\sum_{i=1,1+(n+p),1+2(n+p),\ldots}^{N_{1}-(n+p)+1} s_{i}e_{1}q (s_{i}e_{1}q)'\\
&+\sum_{i=2,2+(n+p), 2+2(n+p),\ldots}^{N_{1}-(n+p)+2} s_{i}e_{2}q(s_{i}e_{2}q)'\\
&+ \cdots\\
&+\sum_{i=n+p,n+p+(n+p),\ldots}^{N_{1}} \ s_{i}e_{n+p}q(s_{i}e_{n+p}q)'\\
&=\sum_{i=1}^{n+p}\sum_{j=1}^{\frac{N_{1}}{n+p}}e_{i}e_{i}'q^2=\sum_{i=1}^{n+p}\frac{N_{1}}{n+p}e_{i}e_{i}'q^2\\
&=\diag(\frac{N_{1}}{n+p}q^2,\cdots,\frac{N_{1}}{n+p}q^2),
\end{aligned}
\end{equation}
where we used the property that $s_{i}^2=1$ for all $i$, and the fact that $N_{1}\bmod{2(n+p)}=0$ for the second equality.

For the second summation after the last equality in \eqref{3 terms lower}, we have
\begin{equation} \label{M2}
\begin{aligned} 
\sum_{i=1}^{N_{1}}m\mathbf{q}_{i}'&=\left(\sum_{i=1,1+(n+p),1+2(n+p),\ldots}^{N_{1}-(n+p)+1}m(s_{i}e_{1}q)'\right)+\cdots\\
&+\left(\sum_{i=n+p,n+p+(n+p),\ldots}^{N_{1}}m(s_{i}e_{n+p}q)'\right)\\
&=\textbf{0}+\textbf{0}+\ldots+\textbf{0}=\textbf{0},
\end{aligned}
\end{equation}
where we used the property that $s_{i}=1$ if $i\in \{j(n+p)+1,j(n+p)+2,\ldots, j(n+p)+(n+p)|j \text{ is even}\}$ and $s_{i}=-1$ if $i\in \{j(n+p)+1,j(n+p)+2,\ldots, j(n+p)+(n+p)|j \text{ is odd}\}$, and the fact that $N_{1}\bmod{2(n+p)}=0$, i.e., the number of positive terms is exactly the same as the number of negative terms for each summation.

Combining the above equalities, we have 
\begin{equation} \label{touse1}
\begin{aligned} 
\lambda_{min}(ZZ')&\geq \lambda_{min}(\diag(\frac{N_{1}}{n+p}q^2,\cdots,\frac{N_{1}}{n+p}q^2))\\
&=\frac{N_{1}}{n+p}q^2. 
\end{aligned}
\end{equation}
Using the property $\floor{\frac{N}{c}}c\geq N-c$ for any $c>0$, we have 
\begin{equation}
\begin{aligned} 
N_{1}=\floor{\frac{N}{2(n+p)}}\times {2(n+p)}\geq N-2(n+p)\geq \frac{N}{2},
\end{aligned}
\end{equation}
where the second inequality is due to our assumption that $N\geq 4(n+p)$.

Hence, the above inequality in conjunction with \eqref{touse1} yields
\begin{equation} 
\begin{aligned} 
\lambda_{min}(ZZ')\geq \frac{Nq^2}{2(n+p)}, 
\end{aligned}
\end{equation}
which is of the desired form.

Next, we prove the upper bound. Denoting $N_{2}=\ceil{\frac{N}{2(n+p)}}\times{2(n+p)}$, using $N\leq N_{2}$, we have
\begin{equation} 
\begin{aligned} 
ZZ'&=\sum_{i=1}^{N}z^{i}_{0}z^{i'}_{0}\preceq \sum_{i=1}^{N_{2}}z^{i}_{0}z^{i'}_{0}\\
&=\sum_{i=1}^{N_{2}}mm'+\sum_{i=1}^{N_{2}}\mathbf{q}_{i}\mathbf{q}_{i}'+\sum_{i=1}^{N_{2}}m\mathbf{q}_{i}'+\sum_{i=1}^{N_{2}}\mathbf{q}_{i}m',
\end{aligned}
\end{equation}
where $z^{1}_{0},z^{2}_{0},\ldots, z^{N_{2}}_{0}$ are generated from Algorithm \ref{algo1} with input parameter $N_{2}$. Since $N_{2}\bmod{2(n+p)}=0$, we can follow a similar procedure as in the proof of the lower bound to obtain $\sum_{i=1}^{N_{2}}m \mathbf{q}_{i}'=\sum_{i=1}^{N_{2}}\mathbf{q}_{i}m'=\textbf{0}$ and $\sum_{i=1}^{N_{2}}\mathbf{q}_{i}\mathbf{q}_{i}'=\diag(\frac{N_{2}}{n+p}q^2,\cdots,\frac{N_{2}}{n+p}q^2)$. Hence, we have
\begin{equation} 
\begin{aligned} 
\lambda_{max}(ZZ')&\leq \lambda_{max}(\sum_{i=1}^{N_{2}}mm'+\sum_{i=1}^{N_{2}}\mathbf{q}_{i}\mathbf{q}_{i}')\\
&\leq N_{2}(\|m\|^2+\frac{q^2}{n+p})\\
&\leq (N+2(n+p))\|m\|^2+(N+2(n+p))\frac{q^2}{n+p}\\
&\leq N(2\|m\|^2+\frac{2q^2}{n+p}),
\end{aligned}
\end{equation}
where the third inequality is due to the relationship $N_{2}\leq N+2(n+p)$, and the last inequality is due to the assumption that $N\geq 4(n+p)$.

\subsection{Proof of Lemma \ref{bound noise}}
Denoting $\bar{V}_{N}=\lambda I_{n+p}+ZZ'$, we have
\begin{equation*}
\begin{aligned}
&\|WZ'(ZZ'+\lambda I_{n+p})^{-1/2}\|=\|\bar{V}_{N}^{-1/2}ZW'\|.
\end{aligned}
\end{equation*}
Let $\hat{V}_{N}=(\lambda+\frac{Nq^2}{2(n+p)})I_{n+p}$. When $N\geq 4(n+p)$, we can apply the lower bound in Lemma \ref{lemma:PE} to get $\bar{V}_{N}\succeq \hat{V}_{N}$. Since $\bar{V}_{N} \succeq \hat{V}_{N} \Rightarrow  2\bar{V}_{N} \succeq \bar{V}_{N}+\hat{V}_{N}\Rightarrow \bar{V}_{N}^{-1} \preceq 2 (\bar{V}_{N}+\hat{V}_{N})^{-1}$, where we used Lemma \ref{inverse bound}, we can write 
\begin{equation*}
\begin{aligned}
&\|\bar{V}_{N}^{-1/2}ZW'\|\leq \sqrt{2}\|(\bar{V}_{N}+\hat{V}_{N})^{-1/2}ZW'\|\\
&=\sqrt{2}\|(\hat{V}_{N}+\lambda I_{n+p}+\sum_{i=1}^{N}z^{i}_{0}z^{i'}_{0})^{-1/2}(\sum_{i=1}^{N}z^{i}_{0}w^{i'}_{0})\|,
\end{aligned}
\end{equation*}
where the inequality is due to Lemma \ref{norm bound}.

Denote $V=\hat{V}_{N}+\lambda I_{n+p}$. Define the filtration $\{\mathcal{F}_{t}\}_{t\geq 0}$, where $\mathcal{F}_{t}=\sigma(\{z_{0}^{i+1}\}_{i=0}^{t}\cup\{w_{0}^{j}\}_{j=1}^{t})$. Since the sequence of $z^{i}_{0}$ generated by Algorithm \ref{algo1} is deterministic, and the noise terms are independent, for any fixed $\delta \in (0,1)$, we can apply Lemma \ref{martingale_bound_multi} to obtain with probability at least $1-\delta$
\begin{equation*}
\begin{aligned}
&\sqrt{2}\|(\bar{V}_{N}+\hat{V}_{N})^{-1/2}ZW'\|\\
&\leq 3 \sigma_{w} \sqrt{\log\frac{9^n}{\delta}+\frac{1}{2}\log\det((V+ZZ')V^{-1})}.
\end{aligned}
\end{equation*}
For the determinant term, we can apply the upper bound in Lemma \ref{lemma:PE} to obtain
\begin{equation*}
\begin{aligned}
det((V+ZZ')V^{-1})&=\frac{\det(V+ZZ')}{\det(V)}\\
&\leq \frac{(2\lambda+\frac{Nq^2}{2(n+p)}+\|ZZ'\|)^{n+p}}{(2\lambda+\frac{Nq^2}{2(n+p)})^{n+p}}\\
&\leq (1+\frac{N(2\|m\|^2+\frac{2q^2}{n+p})}{2\lambda+\frac{Nq^2}{2(n+p)}})^{n+p}\\
&= (1+\frac{4\|m\|^2(n+p)+4q^2}{q^2+\zeta})^{n+p},
\end{aligned}
\end{equation*}
where we used the fact that the determinant is the product of eigenvalues. The result then follows.

\subsection{Proof of Lemma \ref{bound nonlinearity}}
Note that 
\begin{equation} 
\begin{aligned}
\|RZ'(ZZ'+\lambda I_{n+p})^{-1}\|\leq \|R\|\|Z'(ZZ'+\lambda I_{n+p})^{-1}\|.
\end{aligned}
\end{equation}
For the term $\|R\|$, using $R_{i,j}$ to denote its $(i,j)$ entry, we have 
\begin{equation} \label{touse2}
\begin{aligned}
\|R\|\leq \|R\|_{F}=\sqrt{\sum_{i=1}^{n}\sum_{j=1}^{N}R_{i,j}^2}\leq \sqrt{Nn\beta^2  (\|m\|_{1}+q)^4},
\end{aligned}
\end{equation}
where the second inequality is due to the fact that $\|z^{i}_{0}\|_{1}=\|m+\mathbf{q}_{i}\|_{1}\leq \|m\|_{1}+q$ for all $i=1,\ldots,N$, the assumption that $\|m\|_{1}+q< c$, and Assumption \ref{ass:remainder}.

For the term $\|Z'(ZZ'+\lambda I_{n+p})^{-1}\|$, we have
\begin{equation*} 
\begin{aligned}
&\|Z'(ZZ'+\lambda I_{n+p})^{-1}\|\\
&=\sqrt{\|(ZZ'+\lambda I_{n+p})^{-1}ZZ'(ZZ'+\lambda I_{n+p})^{-1}\|}.
\end{aligned}
\end{equation*}
Note that 
\begin{equation} \label{touse3}
\begin{aligned}
&\|(ZZ'+\lambda I_{n+p})^{-1}ZZ'(ZZ'+\lambda I_{n+p})^{-1}\|=\\
&\|(ZZ'+\lambda I_{n+p})^{-1}(ZZ'+\lambda I_{n+p})(ZZ'+\lambda I_{n+p})^{-1}\\
&\quad -\lambda(ZZ'+\lambda I_{n+p})^{-1}(ZZ'+\lambda I_{n+p})^{-1}\|\\
&\leq \|(ZZ'+\lambda I_{n+p})^{-1}\|+\lambda\|(ZZ'+\lambda I_{n+p})^{-1}\|^2.
\end{aligned}
\end{equation}
Furthermore, we have
\begin{equation*} 
\begin{aligned}
\|(ZZ'+\lambda I_{n+p})^{-1}\|&=\frac{1}{\lambda_{min}(ZZ'+\lambda I_{n+p})}\\
&= \frac{1}{\lambda_{min}(ZZ')+\lambda}.
\end{aligned}
\end{equation*}
Using the above inequality and \eqref{touse3}, since $N\geq 4(n+p)$, we can apply Lemma \ref{lemma:PE} to get
\begin{equation*} 
\begin{aligned}
&\|Z'(ZZ'+\lambda I_{n+p})^{-1}\|\\
&\leq \sqrt{\frac{2(n+p)}{Nq^2+2\lambda (n+p)}}+\frac{2\sqrt{\lambda} (n+p)}{Nq^2+2\lambda (n+p)},
\end{aligned}
\end{equation*}
where we used the relationship that $\sqrt{x+y}\leq \sqrt{x}+\sqrt{y}$ for $x,y\geq 0$.

Finally, combining the above inequality with \eqref{touse2}, using $\|m\|_{1}\leq (\sqrt{b}-1)q$, and after some algebraic manipulations, we have the desired result.


\end{document}